\pdfoutput=1
\documentclass[11pt]{article}
\usepackage{xcolor}
\usepackage{ACL2023}
\usepackage{booktabs}
\usepackage{array}
\usepackage{multirow}
\usepackage{makecell}
\usepackage{times}
\usepackage{latexsym}
\usepackage{amsmath}
\usepackage{amssymb}
\usepackage{mathtools}
\usepackage[T1]{fontenc}
\usepackage[most]{tcolorbox}
\usepackage{algorithm}
\usepackage{algorithmic}
\usepackage[utf8]{inputenc}

\usepackage{microtype}

\usepackage{inconsolata}
\usepackage{graphicx}

\title{\emph{FreeChunker}: A Cross-Granularity Chunking Framework}

\author{Wenxuan Zhang\textsuperscript{1}, Yuan-Hao Jiang\textsuperscript{1}, Yang Cao\textsuperscript{1}, Yonghe Wu\textsuperscript{2}\thanks{Corresponding author.} \\
  \textsuperscript{1}Shanghai Institute of Artificial Intelligence for Education, East China Normal University, China \\
  \textsuperscript{2}Education Technology, East China Normal University, China \\
  \texttt{yhwu@deit.ecnu.edu.cn}
}

\begin{document}
\maketitle
\thispagestyle{plain}
\pagestyle{plain}
\begin{abstract}
Chunking strategies significantly impact the effectiveness of Retrieval-Augmented Generation (RAG) systems. Existing methods operate within fixed-granularity paradigms that rely on static boundary identification, limiting their adaptability to diverse query requirements. This paper presents \textbf{\emph{FreeChunker}}, a Cross-Granularity Encoding Framework that fundamentally transforms the traditional chunking paradigm: the framework treats sentences as atomic units and shifts from static chunk segmentation to flexible retrieval supporting arbitrary sentence combinations. This paradigm shift not only significantly avoids the computational overhead required for semantic boundary detection, but also enhances adaptability to complex queries. Experimental evaluation on LongBench V2 demonstrates that \textbf{\emph{FreeChunker}} possesses significant advantages in both retrieval performance and time efficiency compared to existing chunking methods. The pre-trained models and codes are available at \url{https://github.com/mazehart/FreeChunker}.
\end{abstract}

\section{Introduction}

The rapid advancement of large language models (LLMs) has substantially propelled progress in natural language processing, enabling strong generalization capabilities across diverse tasks~\cite{DBLP:conf/nips/BrownMRSKDNSSAA20}. However, LLMs typically rely on static, parameterized knowledge, often making them less suitable for scenarios requiring up-to-date or domain-specific information. This limitation frequently leads to hallucinated content that may deviate from factual correctness~\cite{huang2025survey, DBLP:conf/emnlp/LiCZNW23}.

To address this limitation, RAG has emerged as a promising paradigm. By incorporating external knowledge sources, RAG enhances the generation process with more timely and verifiable content, achieving strong performance on knowledge-intensive tasks such as open-domain question answering and fact verification~\cite{lewis2020retrieval}. However, the effectiveness of RAG critically depends on the quality of retrieved content, which is substantially influenced by document chunking strategies~\cite{DBLP:conf/nips/RuQHZSCJWSLZWJ024}. Prior research has demonstrated that chunking granularity often directly affects intra-chunk coherence, semantic coverage, and downstream generation quality. Inappropriate chunking strategies may disrupt semantic structures, potentially leading to information fragmentation or redundancy, ultimately undermining both retrieval and generation performance~\cite{bradland2025hope,lyu2025crud}.

\begin{figure}[t]
\centering 
\includegraphics[width=0.49\textwidth]{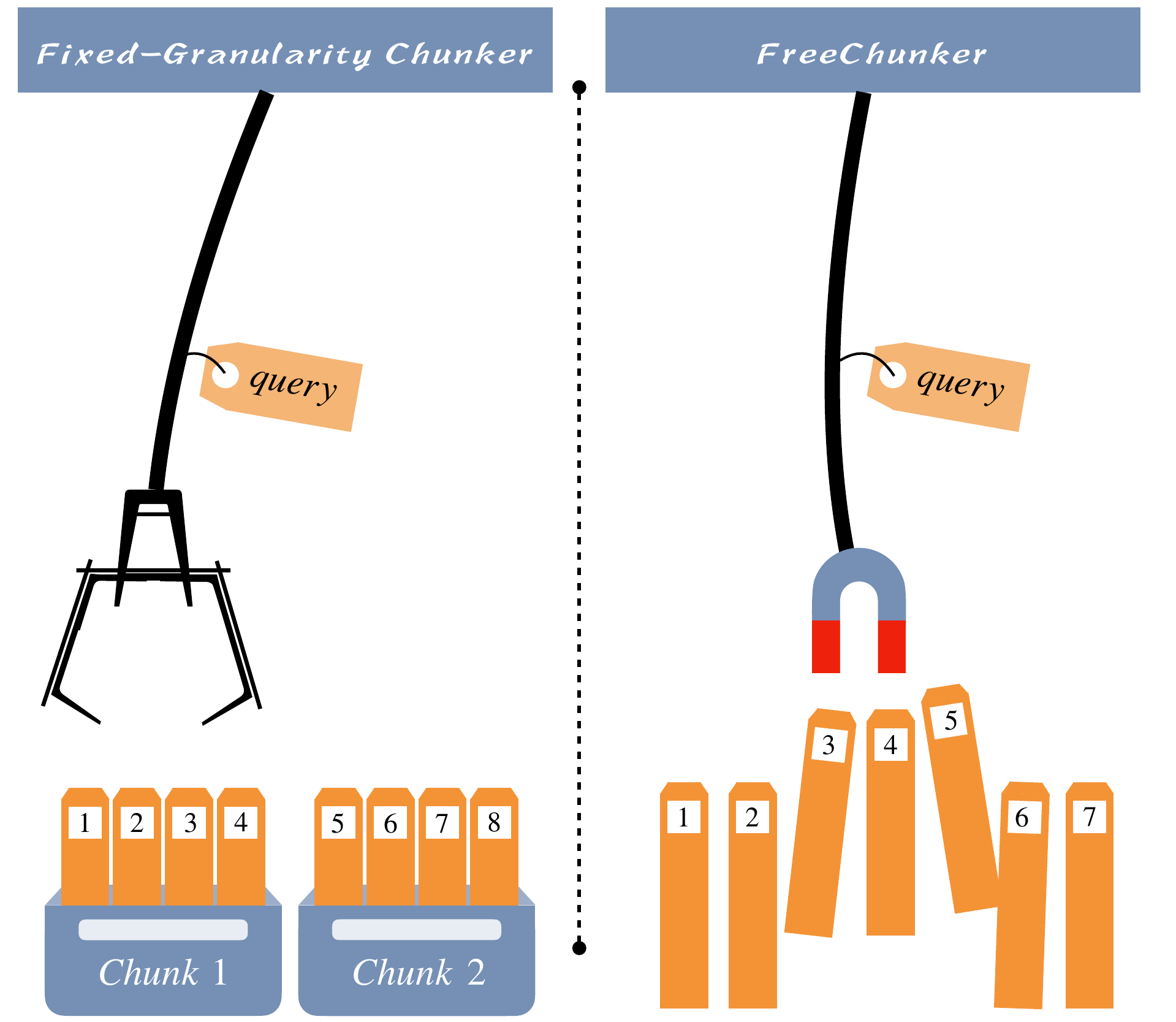}
\caption{\emph{FreeChunker} breaks the limitation of fixed-granularity text chunking, enabling flexible extraction of sentence combinations at arbitrary granularities.}
\label{fig:freechunker_overview}
\end{figure}

To improve chunking quality, recent efforts have explored semantic-aware chunking techniques. For instance,  SemanticChunker~\cite{langchain2024semanticchunker} segments documents based on sentence embedding similarity. Meta-Chunking~\cite{DBLP:journals/corr/abs-2410-12788} identifies semantic boundaries through perplexity analysis and probability-based segmentation decisions. LumberChunker~\cite{DBLP:conf/emnlp/DuarteMGF0O24} leverages LLMs to predict chunk boundaries more aligned with human interpretation. However, existing approaches often lack the flexibility to adapt to diverse query intents or varying deployment constraints. Moreover, some methods rely on computationally expensive large models, where the marginal gains of semantic chunking may not always justify the computational overhead~\cite{DBLP:conf/naacl/QuTB25}. Structurally, these approaches still operate within a single-granularity paradigm, consequently limiting their ability to simultaneously accommodate different information needs within the same document collection. Furthermore, the reliance on complex model inference for boundary detection introduces severe latency, rendering them impractical for large-scale real-time retrieval.

To overcome the intrinsic constraints of fixed chunking schemes and the computational inefficiency of existing semantic approaches, \textbf{\emph{FreeChunker}} is proposed in this paper—a novel Cross-Granularity Encoding Framework that introduces an alternative chunking paradigm. As illustrated in Figure~\ref{fig:freechunker_overview}, unlike traditional chunking methods that are typically constrained by fixed text boundaries, the proposed framework enables retrievers to extract arbitrary combinations of sentences from documents, thereby providing enhanced flexibility in content retrieval. The core contributions of this paper are summarized as follows:

\begin{itemize}
    \item \textbf{Cross-granularity chunking paradigm}: A novel approach is proposed that breaks the limitation of fixed-granularity text chunking, enabling flexible extraction of sentence combinations at arbitrary granularities.
    
    \item \textbf{Computational efficiency}: The framework treats sentences as atomic units, avoiding time-consuming boundary identification processes. Instead, it utilizes preset granularity combinations, significantly reducing both chunking and encoding time.
    
    \item \textbf{Superior performance}: The framework achieves the best average retrieval performance by adapting multiple granularity levels to diverse query requirements, consistently outperforming existing baselines across various scenarios on the LongBench V2 benchmark.

\end{itemize}

\begin{figure*}[t]
    \centering
    \includegraphics[width=\linewidth]{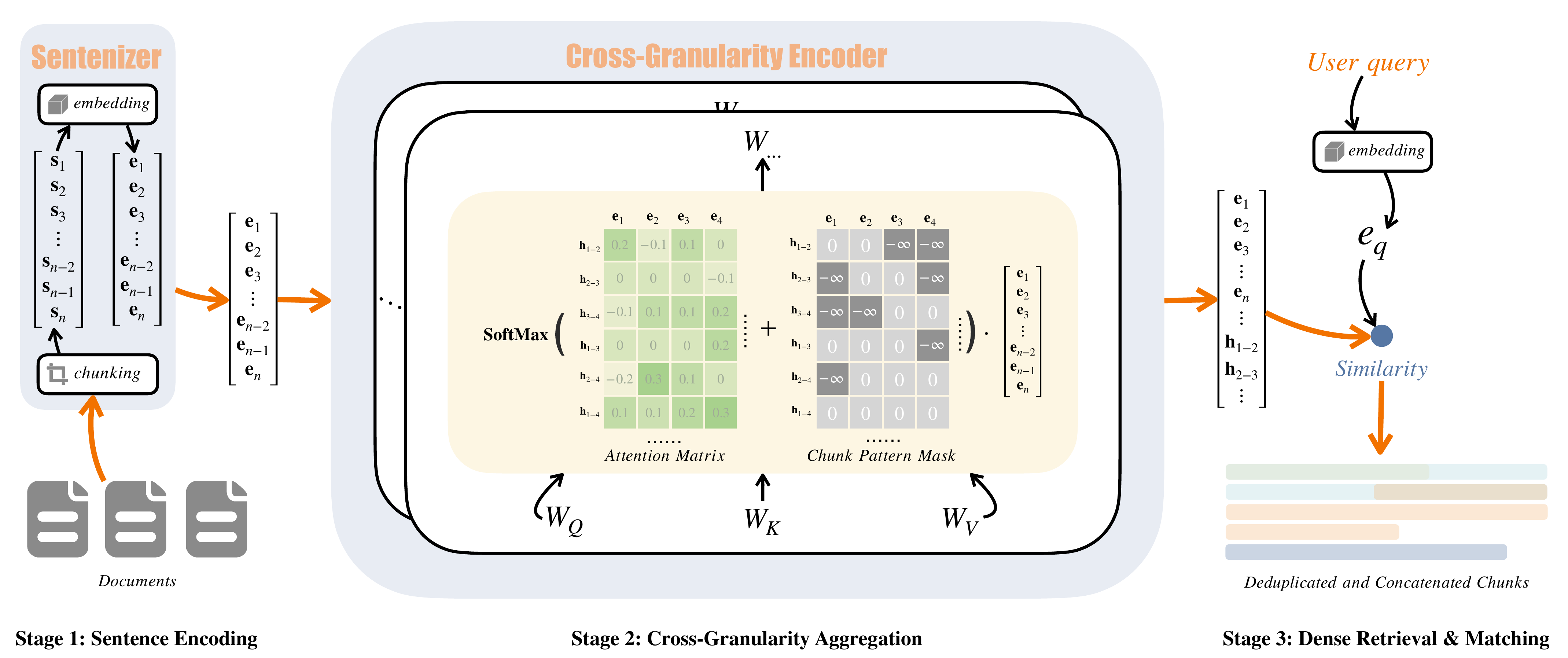}
    \caption{\textbf{The architecture of \emph{FreeChunker}}. Long text input is processed by the Sentenizer and Embedding Model to produce sentence embeddings. The Attention Matrix and Chunk Pattern Mask work together to generate multiple granularity chunk embeddings in a single forward pass, enabling flexible on-demand access to different chunk sizes.}
    \label{fig:architecture}
\end{figure*}
\section{Related Work}

\paragraph{Limitations of Fixed-granularity Chunking}
As a foundational step in RAG, text chunking substantially influences overall system performance. The most straightforward strategy divides documents into fixed-granularity chunks, which is simple to implement but often misaligns with semantic boundaries, consequently leading to information fragmentation or redundancy~\cite{DBLP:conf/nips/RuQHZSCJWSLZWJ024}. More critically, fixed chunking forces a single granularity choice across the entire document, thereby preventing the system from simultaneously accessing both fine-grained details and coarse-grained context that different queries may require. This granularity rigidity fundamentally limits the retrieval system's ability to adapt to diverse information needs within a single document collection.

\paragraph{Semantic-aware Chunking Techniques}
To address the aforementioned semantic boundary issues, recent work has explored more adaptive chunking strategies. SemanticChunker~\cite{langchain2024semanticchunker} relies on sentence-level embedding similarity to enhance intra-chunk semantic coherence by identifying natural breakpoints based on semantic similarity thresholds. Meta-Chunking~\cite{DBLP:journals/corr/abs-2410-12788} proposes two adaptive segmentation algorithms leveraging large language models' logical awareness: Perplexity Chunking identifies logical boundaries by analyzing inter-sentence perplexity variations, while Margin Sampling Chunking determines boundaries by comparing model probability differences for segmentation decisions. LumberChunker~\cite{DBLP:conf/emnlp/DuarteMGF0O24} leverages large language models to determine chunk boundaries more aligned with human interpretation. However, despite their semantic awareness, these methods still operate under a single granularity paradigm—each document is ultimately represented at one chosen level of granularity, whether sentence-level, paragraph-level, or custom semantic units.

\paragraph{Adaptive Attempts}
Inspired by the concept of mixture of experts (MoE)~\cite{jacobs1991adaptive, DBLP:conf/iclr/ShazeerMMDLHD17}, recent methods like MoC~\cite{DBLP:journals/corr/abs-2503-09600} and MoG (Mix-of-Granularity)~\cite{DBLP:conf/coling/ZhongLCZQ25} attempt to enhance adaptability by treating different chunking strategies as expert modules and employing routing mechanisms to dynamically select the most suitable chunker based on query characteristics. However, these MoE-inspired methods essentially aggregate existing chunkers as experts; consequently, both the quality of semantic chunking and the degree of granularity control remain fundamentally bounded by the capabilities of the underlying chunker experts.

Despite these advances, a fundamental limitation persists: current methods still operate within a single-chunking, single-granularity paradigm—whether fixed, semantics-driven, or query-adaptive. Consequently, retrieval systems cannot simultaneously access multiple levels of information granularity, which is precisely what is needed to better accommodate diverse query intents. Therefore, there remains a critical need for a paradigm that can transcend these single-granularity constraints.

\section{Methodology}
\label{sec:method}

In \emph{FreeChunker}, sentences are treated as atomic units and chunk embeddings are generated across multiple granularities simultaneously. This approach enables retrieval systems to access any desired chunk size on demand, ranging from fine-grained single sentences to coarse-grained multi-sentence contexts.

Before detailing the mathematical formulation, a high-level intuition of the approach is provided. Unlike traditional methods that physically segment text and encode each chunk separately, \emph{FreeChunker} employs a unified encoding process. Imagine a ``mask'' overlaid on the document's sentence representations. By adjusting this mask, it is possible to dynamically specify which sentences should be attended to and combined into a chunk embedding. Crucially, by constructing a composite mask that contains patterns for multiple granularities (e.g., 1-sentence, 2-sentence, ..., k-sentence chunks), embeddings for all desired chunk sizes can be generated in a single forward pass of the model. This eliminates the need for redundant computations and allows for flexible, overlapping chunk definitions without repeated encoding.

\subsection{Sentenizer}

The \textbf{Sentenizer} serves as the initial stage of the \emph{FreeChunker} pipeline. Given a document $\mathcal{D}$, it first decomposes it into atomic sentence units $\mathcal{S} = \{s_1, s_2, \dots, s_n\}$. Subsequently, a token-level embedding model $\mathcal{M}$ is employed to encode these sentences into a sentence embedding matrix $\mathcal{E} = [\vec{e_1}, \vec{e_2}, \dots, \vec{e_n}] \in \mathbb{R}^{n \times d}$, where $d$ denotes the embedding dimension. This process provides the foundational input for the subsequent cross-granularity encoding process.

\subsection{Cross-Granularity Chunk Pattern}

The core innovation of \emph{FreeChunker} lies in proposing a flexible multi-granularity paradigm that overcomes the fixed granularity limitation of existing methods by defining multiple chunk patterns, thereby enabling on-demand granularity access.

Based on the $n$ sentences obtained from the Sentenizer, a Chunk Pattern Mask $\mathbf{P} \in \{0, -\infty\}^{m \times n}$ is constructed, where $m$ is the number of desired chunks. Each row $\mathbf{P}[i,:]$ represents the $i$-th chunk pattern, and $\mathbf{P}[i,j] = 0$ indicates that sentence $s_j$ belongs to that pattern.

The key insight is that the matrix can encode arbitrary granularity combinations. For any granularity $g$ and starting position $s$, chunk patterns are defined as:
\begin{equation}
\mathbf{P}_{g,s}[i,j] = \begin{cases}
0 & \text{if } s \leq j < s + g \\
-\infty & \text{otherwise}
\end{cases}
\end{equation}

\begin{figure*}[htbp]
\centering
\includegraphics[width=\textwidth]{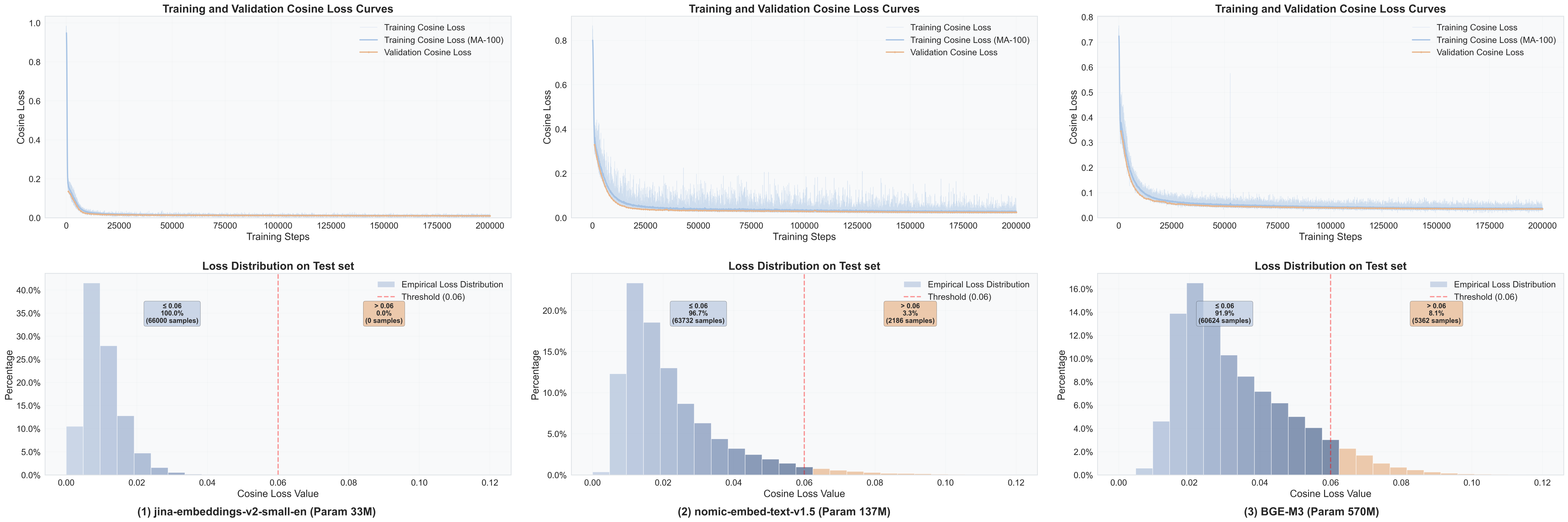}
\caption{\textbf{Training Dynamics and Generalization Across Embedding Models.} Training and validation curves (top) illustrate training loss in blue and validation loss in orange, and test set loss distributions (bottom) show the cosine loss on an in-distribution test set, demonstrating consistent convergence and effective generalization.}
\label{fig:training_curves}
\end{figure*}

Therefore, the framework supports the superposition processing of different granularities. As illustrated in Stage 2 of Figure~\ref{fig:architecture}, this results in a mask with a distinctive banded structure, where valid attention weights (0) are clustered around the diagonal for adjacent sentence chunks, while other areas are masked out ($-\infty$).

\subsection{Cross-Granularity Encoder}

After obtaining the Chunk Pattern Mask $\mathbf{P}$ and sentence embeddings $\mathcal{E}$, independently encoding each chunk would typically cause extensive redundant computation. For any chunk spanning from sentence $s_i$ to sentence $s_j$, its true embedding is conceptually equivalent to $\vec{e}_{i,j} = \mathcal{M}([s_i, s_{i+1}, \dots, s_j])$, where $[s_i, s_{i+1}, \dots, s_j]$ represents the concatenation of sentences.

To efficiently compute these representations, a specialized transformer architecture is modified. For each layer, a learnable chunk embedding $\mathbf{h}_{\text{chk}} \in \mathbb{R}^{d}$ is introduced that is replicated $m$ times to form $\mathcal{H} = [\mathbf{h}_{\text{chk}}, \mathbf{h}_{\text{chk}}, \dots, \mathbf{h}_{\text{chk}}]^T \in \mathbb{R}^{m \times d}$. The cross-granularity attention is then computed using the sentence embeddings $\mathcal{E}$ as Keys and Values:
\begin{equation}
\mathbf{Q} = \mathbf{W}_Q \mathcal{H}, \quad \mathbf{K} = \mathbf{W}_K \mathcal{E}, \quad \mathbf{V} = \mathbf{W}_V \mathcal{E}
\end{equation}
\begin{equation}
\text{Attn}(\mathcal{H}, \mathcal{E}) = \mathrm{softmax}\!\Big(\tfrac{\mathbf{Q}\mathbf{K}^\top}{\sqrt{d}} + \mathbf{P}\Big) \mathbf{V}
\end{equation}

Following the attention mechanism, standard Transformer layer operations are applied, including Layer Normalization and Feed-Forward Networks (FFN), to produce the final chunk embeddings. This design allows parallel generation of all chunk embeddings in a single forward pass, with sentence-level embeddings being reused across different granularities, thereby avoiding redundant encoding of sentences.

\subsection{Deduplicate and Concatenate}
\label{sec:dedup_concat}

To address the overlap in multi-granularity retrieved chunks, a post-processing strategy is adopted. Overlapping segments are deduplicated by marking the starting positions, and the retained unique segments are concatenated in order to reconstruct the narrative context.

\section{Training Process}
\label{sec:training}

\subsection{Training Setup}

\paragraph{Token-level Embedding Models}
To train the cross-granularity framework (Section~\ref{sec:method}) for effective chunk representation, three fundamental token-level embedding models are selected: jina-embeddings-v2-small-en (33M parameters), nomic-embed-text-v1.5 (137M parameters), and BGE-M3 (570M parameters). These models represent different scales and architectures, enabling a thorough assessment across varying embedding capacities.

\paragraph{Dataset and Preprocessing}
The training dataset comprises 100K documents sampled from The Pile~\cite{gao2020pile}, each truncated to 8000 tokens and split into sentences. Training pairs $(\vec{e}, \vec{v})$ are constructed across 5 granularities (2, 4, 8, 16, 32 sentences), where $\vec{e}$ is the ground-truth embedding obtained by encoding concatenated sentences using the base models, and $\vec{v}$ denotes the corresponding output from the proposed framework.

\paragraph{Optimization}
The training objective is formulated to minimize the cosine similarity loss:
\begin{equation}
\mathcal{L} = 1 - \frac{1}{|\mathcal{B}|} \sum_{(\vec{e}, \vec{v}) \in \mathcal{B}} \cos(\vec{e}, \vec{v})
\end{equation}
where $\mathcal{B}$ denotes the training batch. The AdamW optimizer is employed with learning rate $1 \times 10^{-4}$ and batch size 1. Training is conducted for 2 epochs with cosine schedule learning rate decay, incorporating warmup over the first one-third of total training steps. Model validation is performed every 1000 steps on a held-out validation set of 200 samples to monitor generalization performance.

\paragraph{Computational Resources}
All training experiments are conducted on NVIDIA A800 GPUs with 80GB memory. The total computational cost exceeds 500 GPU hours, encompassing dataset encoding construction and model training processes. 

\subsection{Training Dynamics and Evaluation}

\paragraph{Convergence Analysis}
The upper panels of Figure~\ref{fig:training_curves} present the training and validation cosine loss curves for three base models. All configurations demonstrate smooth and stable convergence within 200K steps, validating the effectiveness of the overall optimization strategy. Notably, the smallest model, jina-embeddings-v2-small-en (33M parameters), exhibits the fastest and cleanest convergence with lower training loss. As model parameters increase, both training and validation losses show degradation across the three models. This phenomenon may be attributed to the use of a fixed 330M-parameter sentence-level encoder for all three base models, suggesting that scaling the sentence-level encoder capacity could potentially improve fitting performance for BGE-M3 (570M parameters).

\paragraph{Test Set Generalization Analysis}
The lower panels of Figure~\ref{fig:training_curves} illustrate the cosine loss distributions on the independent test set. A threshold of 0.06 is employed to distinguish well-fitted samples from under-fitted ones, enabling coarse-grained assessment of chunk embedding accuracy. For jina-embeddings-v2-small-en, 100\% of test samples fall below this threshold, with loss distribution highly concentrated in the 0.01–0.02 range, reflecting excellent generalization capability and minimal approximation error. In contrast, nomic-embed-text-v1.5 (137M parameters) exhibits a more dispersed distribution: 96.7\% of samples remain below the 0.06 threshold, yet long-tail high-loss samples emerge, indicating increased representation noise or model uncertainty. BGE-M3 (570M parameters) shows further distribution broadening, consistent with the training loss patterns observed.

\section{Theoretical Analysis}
\label{sec:analysis}

While Section~\ref{sec:training} demonstrates that the cross-granularity encoder achieves a low fitting loss (i.e., high cosine similarity $\rho \approx 1$ between true and approximate embeddings), it remains crucial to understand how this approximation error propagates to the downstream retrieval task.

\subsection{Setup and Constraints}
Let $\vec{e}_{i,j}=\mathcal{M}([s_i, \dots, s_j])$ denote the \emph{true} chunk embeddings of the concatenated chunk $[s_i, \dots, s_j]$, and let $\vec{v}_{i,j}$ be the \emph{approximate} chunk embeddings produced by the cross-granularity encoder. Let $\vec{q}$ be a query embedding. Throughout this section all vectors are assumed to be unit-norm.

The substitution loss is studied as follows:
\begin{equation}
\varepsilon \;:=\; \big|\cos(\vec{q},\vec{e}_{i,j})-\cos(\vec{q},\vec{v}_{i,j})\big|.
\end{equation}
Two parameters characterize the geometric configuration:
\begin{equation}
\label{eq:constraints}
\begin{aligned}
\rho &\;=\; \cos(\vec{e}_{i,j},\vec{v}_{i,j}), \\
s &\;=\; \cos(\vec{q},\vec{e}_{i,j}),
\end{aligned}
\end{equation}
where $\rho$ quantifies substitution quality and $s$ captures query alignment with the true chunk.

\subsection{Upper Bound Analysis}

To rigorously quantify the retrieval quality in high-dimensional embedding spaces (e.g., $d \ge 768$), a probabilistic bound is established by leveraging the \textit{Concentration of Measure} phenomenon. This derivation proves that the probability of a random query aligning with a specific approximation error direction is exponentially small. 

\paragraph{Step 1: Orthogonal Decomposition}
A coordinate system is established aligned with the true chunk embedding $\vec{e}_{i,j}$. Any unit vector $\vec{x}$ can be decomposed into a component parallel to $\vec{e}_{i,j}$ and a component in the orthogonal complement $\vec{e}_{i,j}^\perp$: $\vec{x} = (\vec{x} \cdot \vec{e}_{i,j})\vec{e}_{i,j} + \vec{x}_{\perp}$.

Given $\vec{v}_{i,j} \cdot \vec{e}_{i,j} = \rho$, the magnitude of the orthogonal component is $\sqrt{1-\rho^2}$. It can be written as:
\begin{equation}
\vec{v}_{i,j} = \rho \vec{e}_{i,j} + \sqrt{1-\rho^2} \, \mathbf{u},
\end{equation}
where $\mathbf{u} \in \mathbb{R}^d$ is a unit vector in the orthogonal subspace ($\mathbf{u} \perp \vec{e}_{i,j}$). Intuitively, $\mathbf{u}$ represents the \emph{direction} of the approximation error.

Given $\vec{q} \cdot \vec{e}_{i,j} = s$, the query is similarly decomposed:
\begin{equation}
\vec{q} = s \vec{e}_{i,j} + \sqrt{1-s^2} \, \mathbf{w},
\end{equation}
where $\mathbf{w} \in \mathbb{R}^d$ is a unit vector in the orthogonal subspace ($\mathbf{w} \perp \vec{e}_{i,j}$), determined solely by the query.

\paragraph{Step 2: Algebraic Expansion of Loss}
Substituting the decomposed forms into the dot product $\vec{q} \cdot \vec{v}_{i,j}$:
\begin{align}
\vec{q} \cdot \vec{v}_{i,j} &= (s \vec{e}_{i,j} + \sqrt{1-s^2} \mathbf{w}) \cdot (\rho \vec{e}_{i,j} + \sqrt{1-\rho^2} \mathbf{u}) \nonumber \\
&= s\rho (\vec{e}_{i,j} \cdot \vec{e}_{i,j}) + \sqrt{1-s^2}\sqrt{1-\rho^2}(\mathbf{w} \cdot \mathbf{u}) \nonumber \\
&= s\rho + \sqrt{1-s^2}\sqrt{1-\rho^2}(\mathbf{w} \cdot \mathbf{u}).
\end{align}
Note that cross-terms vanish due to orthogonality. Substituting this back into the loss expression $\varepsilon = |s - \vec{q} \cdot \vec{v}_{i,j}|$:
\begin{align}
\varepsilon &= \left| s - \left( s\rho + \sqrt{1-s^2}\sqrt{1-\rho^2}(\mathbf{w} \cdot \mathbf{u}) \right) \right| \nonumber \\
&= \left| s(1-\rho) - \sqrt{1-s^2}\sqrt{1-\rho^2}(\mathbf{w} \cdot \mathbf{u}) \right|.
\end{align}
Using the triangle inequality $|a-b| \le |a| + |b|$, the structural upper bound is obtained:
\begin{equation}
\label{eq:structural_bound}
\varepsilon \le \underbrace{s(1-\rho)}_{\text{Radial Error}} + \sqrt{1-s^2}\sqrt{1-\rho^2} \underbrace{|\mathbf{w} \cdot \mathbf{u}|}_{\text{Tangential Noise}}.
\end{equation}

\paragraph{Step 3: Probabilistic Bound via Concentration}
The term $|\mathbf{w} \cdot \mathbf{u}|$ represents the projection of the error direction $\mathbf{u}$ onto the query's specific orthogonal direction $\mathbf{w}$. Both vectors reside in a $(d-1)$-dimensional subspace.

\noindent\textbf{Assumption (Quasi-Random Error).} For an independent query, the direction of the approximation error $\mathbf{u}$ is uniformly distributed on the sphere $S^{d-2}$ relative to $\mathbf{w}$.

While $\mathbf{u}$ arises from aggregation encoding, it is generally decoupled from the semantic intent of user queries. In high-dimensional manifolds ($d \gg 100$), the probability of an independent query aligning with a specific structural artifact vanishes, rendering them effectively orthogonal.

To quantify this, \textbf{Levy's Lemma}~\citep{milman1986asymptotic,ledoux2001concentration} is applied. For a random vector $\mathbf{u}$ uniformly distributed on the sphere and a fixed vector $\mathbf{w}$, the probability that their projection magnitude $|\mathbf{w} \cdot \mathbf{u}|$ exceeds a threshold $t$ decays exponentially:
\begin{equation}
P(|\mathbf{w} \cdot \mathbf{u}| \ge t) \le 2 \exp\left(-\frac{(d-1) t^2}{2}\right).
\end{equation}
Let $\delta \in (0,1)$ be a small failure probability. By equating the tail probability bound to $\delta$, the threshold $t$ is derived as follows:
\begin{equation}
\begin{split}
2 \exp\left(-\frac{(d-1) t^2}{2}\right) &= \delta \\
\Longrightarrow \quad t &= \sqrt{\frac{2 \ln(2/\delta)}{d-1}}.
\end{split}
\end{equation}
This derivation implies that the projection error $|\mathbf{w} \cdot \mathbf{u}|$ is confined within this threshold $t$ with a confidence level of at least $1-\delta$.

\noindent\textbf{Theorem (Probabilistic Substitution Bound).} 
With probability at least $1-\delta$, the substitution loss is bounded by:
\begin{equation}
\varepsilon \le s(1-\rho) + \sqrt{1-s^2}\sqrt{1-\rho^2} \cdot \sqrt{\frac{2 \ln(2/\delta)}{d-1}}.
\end{equation}

\paragraph{Implications.}
This result highlights the ``blessing of dimensionality'': as $d$ increases (e.g., $d \to \infty$), the tangential noise term $O(1/\sqrt{d})$ vanishes. The substitution loss is thus dominated by the radial term $s(1-\rho)$. Given that the query similarity $s$ is typically less than 1 ($s < 1$) and the fitting process ensures $\rho$ is very close to 1 ($\rho \approx 1$), the term $s(1-\rho)$ becomes negligible. This confirms that the retrieval quality of the fitted vectors is theoretically guaranteed, verifying the robustness of \emph{FreeChunker} in high-dimensional embedding spaces.

\begin{table*}[!ht]
\centering
\resizebox{\textwidth}{!}{%
\small
\setlength{\tabcolsep}{1.5pt}
\begin{tabular}{l|>{\centering\arraybackslash}p{1.2cm}>{\centering\arraybackslash}p{1.2cm}|>{\centering\arraybackslash}p{1.2cm}>{\centering\arraybackslash}p{1.2cm}|>{\centering\arraybackslash}p{1.2cm}>{\centering\arraybackslash}p{1.2cm}|>{\centering\arraybackslash}p{1.2cm}>{\centering\arraybackslash}p{1.2cm}|>{\centering\arraybackslash}p{1.2cm}>{\centering\arraybackslash}p{1.2cm}|>{\centering\arraybackslash}p{1.2cm}>{\centering\arraybackslash}p{1.2cm}}
\toprule
\multirow{2}{*}{Embeddings} & \multicolumn{2}{c|}{\small\textbf{Traditional}} & \multicolumn{2}{c|}{\small\textbf{SemanticChunker}} & \multicolumn{2}{c|}{\small\textbf{PPL Chunking}} & \multicolumn{2}{c|}{\small\textbf{Margin Sampling}} & \multicolumn{2}{c|}{\small\textbf{LumberChunker}} & \multicolumn{2}{c}{\small\textbf{\emph{FreeChunker}}} \\
\cmidrule(lr){2-3} \cmidrule(lr){4-5} \cmidrule(lr){6-7} \cmidrule(lr){8-9} \cmidrule(lr){10-11} \cmidrule(lr){12-13}
& Top-5 & Top-10 & Top-5 & Top-10 & Top-5 & Top-10 & Top-5 & Top-10 & Top-5 & Top-10 & Top-5 & Top-10 \\
\midrule
\multicolumn{13}{c}{\textbf{I. Single-Document QA}} \\[-0.4em]
\midrule
\textbf{jina-small-en}& 30.86& 35.43& 33.71& 33.71& 34.10& 32.57& 28.57& 34.29& 32.00& 35.43& \textbf{35.43}& 32.57 \\
\textbf{nomic-text-v1.5}& 29.90& 29.71& 29.71& 31.43& 32.19& 33.71& 32.57& 31.81& 29.52& 29.14& \textbf{35.62}& \underline{32.57} \\
\textbf{BGE-M3}& 33.33& 37.71& 32.57& 31.43& 33.33& 34.29& 32.57& 33.14& 35.81& 36.38& \textbf{36.57}& 34.29 \\
\midrule
\multicolumn{13}{c}{\textbf{II. Multi-Document QA}} \\[-0.4em]
\midrule
\textbf{jina-small-en}& 24.80& 28.00& 28.80& 28.80& 28.00& 28.27& 29.60& 27.20& 25.87& 28.80& 26.40& 26.40 \\
\textbf{nomic-text-v1.5}& 26.40& 26.40& 22.40& 24.00& 25.07& 27.20& 28.00& 26.40& 25.60& 28.80& 24.00& \underline{28.00} \\
\textbf{BGE-M3}& 27.20& 28.80& 25.87& 31.20& 27.20& 28.00& 27.20& 29.60& 28.80& 29.60& \underline{28.00}& \underline{29.60} \\
\midrule
\multicolumn{13}{c}{\textbf{III. Code Repository Understanding}} \\[-0.4em]
\midrule
\textbf{jina-small-en}& 44.00& 51.33& 36.00& 44.00& 38.00& 48.00& 42.00& 48.00& 44.00& 52.67& \textbf{46.00}& 44.00 \\
\textbf{nomic-text-v1.5}& 42.00& 46.00& 38.00& 36.00& 35.33& 38.00& 44.00& 42.00& 32.00& 36.00& 34.00& \underline{42.00} \\
\textbf{BGE-M3}& 42.00& 44.00& 44.00& 36.67& 48.00& 46.00& 40.00& 36.00& 40.67& 38.00& 42.00& \textbf{48.00} \\
\midrule
\multicolumn{13}{c}{\textbf{IV. Long In-context Learning}} \\[-0.4em]
\midrule
\textbf{jina-small-en}& 27.57& 30.86& 29.63& 29.63& 28.40& 28.40& 25.93& 20.99& 32.10& 29.63& \textbf{32.10}& 28.40 \\
\textbf{nomic-text-v1.5}& 28.40& 28.40& 28.40& 29.63& 26.75& 19.75& 26.75& 22.22& 22.63& 29.63& 20.99& 24.69 \\
\textbf{BGE-M3}& 26.75& 25.10& 34.57& 30.86& 27.57& 28.40& 32.10& 30.45& 28.40& 25.93& 27.16& 25.93 \\
\midrule
\multicolumn{13}{c}{\textbf{V. Long-dialogue History Understanding}} \\[-0.4em]
\midrule
\textbf{jina-small-en}& 26.50& 33.33& 25.64& 17.95& 23.08& 16.24& 28.21& 25.64& 30.77& 23.08& \underline{28.21}& 23.08 \\
\textbf{nomic-text-v1.5}& 35.04& 25.64& 30.77& 20.51& 20.51& 17.95& 23.08& 33.33& 23.08& 25.64& \textbf{38.46}& \textbf{38.46} \\
\textbf{BGE-M3}& 23.08& 23.08& 33.33& 29.06& 14.53& 30.77& 25.64& 37.61& 25.64& 20.51& \textbf{33.33}& \underline{33.33} \\
\midrule
\multicolumn{13}{c}{\textbf{VI. Long Structured Data Understanding}} \\[-0.4em]
\midrule
\textbf{jina-small-en}& 15.15& 24.24& 25.25& 29.29& 33.33& 33.33& 30.30& 36.36& 26.26& 21.21& \textbf{42.42}& \textbf{45.45} \\
\textbf{nomic-text-v1.5}& 35.35& 30.30& 27.27& 21.21& 33.33& 37.37& 39.39& 39.39& 12.12& 12.12& \underline{36.36}& 33.33 \\
\textbf{BGE-M3}& 30.30& 38.38& 23.23& 27.27& 33.33& 33.33& 33.33& 30.30& 21.21& 16.16& \textbf{45.45}& \textbf{42.42} \\
\midrule
\multicolumn{13}{c}{\textbf{VII. Overall}} \\[-0.4em]
\midrule
\textbf{jina-small-en}& 28.76& 33.53& 30.88& 31.34& 31.15& 31.15& 29.82& 31.21& 31.21& 32.67& \textbf{33.60}& 31.61 \\
\textbf{nomic-text-v1.5}& 30.75& 30.02& 28.43& 28.23& 29.03& 29.29& 31.34& 30.55& 26.04& 28.43& 30.48& \textbf{31.61} \\
\textbf{BGE-M3}& 30.62& 33.00& 31.81& 31.35& 30.88& 32.61& 31.41& 32.27& 31.61& 30.62& \textbf{33.80}& \textbf{33.60} \\
\bottomrule
\textbf{Average}& 30.04& \underline{32.18}& 30.37& 30.31& 30.35& 31.01& \underline{30.86}& 31.35& 29.62& 30.57& \textbf{32.63}& \textbf{32.27} \\
\bottomrule
\end{tabular}
}
\caption{\textbf{Average accuracy (\%) of different chunking methods on LongBench V2.} "Overall" denotes the average accuracy across all questions. Standard deviations (SDs) across 3 runs are minimal: 78.7\% of accuracy SDs are 0.00\%. Specifically, FreeChunker show all accuracy SDs $<$ 0.50\%, while others range up to 3.50\%. The best and second-best results for \textbf{\emph{FreeChunker}} and the \textbf{Average} score are highlighted in \textbf{bold} and \underline{underlined}, respectively.}
\label{tab:longbench_chunking}
\end{table*}
\section{Experiments}

\subsection{Experimental Setup}

\paragraph{Datasets}
Comprehensive experiments are conducted on LongBench V2~\cite{bai2024longbench2}, a challenging benchmark specifically designed for long-context understanding tasks. This dataset is suitable for evaluating chunking methods as it contains documents with extensive contexts and covers diverse task types with varying difficulty levels, thereby providing a realistic testbed for assessing the effectiveness of different chunking strategies in practical RAG scenarios.

\paragraph{Embedding Models}
To ensure fair comparison, identical text encoding models are utilized across all methods. Three embedding models are employed for comprehensive evaluation: jina-embeddings-v2-small-en (33M parameters), nomic-embed-text-v1.5 (137M parameters), and BGE-M3 (570M parameters). These models typically represent different scales and architectures, thereby enabling thorough assessment of chunking performance across varying embedding capacities.

\paragraph{Generative Model}
For text generation, Qwen3-8B~\cite{qwen3technicalreport} is employed as the backbone language model, deployed via vLLM~\cite{kwon2023efficient} for efficient inference. To guarantee fair comparison and reproducibility, greedy decoding with temperature set to 0 is  employed by Qwen3-8B.

\begin{figure*}[t]
\centering
\includegraphics[width=\textwidth]{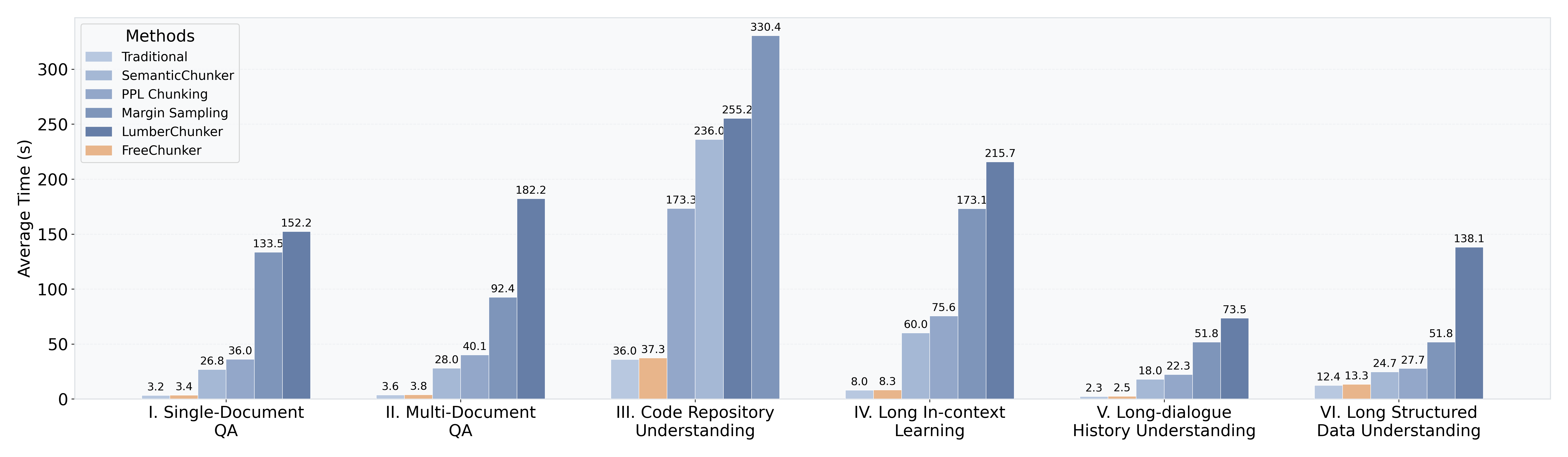}
\caption{\textbf{Average processing time per instance across baselines on LongBench V2.} The processing time comprises the sum of chunking and encoding durations. \emph{FreeChunker} (highlighted in orange) achieves comparable speed to Traditional Chunking and significantly outperforms other methods.}
\label{fig:time_efficiency}
\end{figure*}

\paragraph{Baseline Methods}
\emph{FreeChunker} is compared against several representative chunking approaches. \textbf{(1) Traditional Chunking} represents the traditional approach that first splits text by periods to preserve semantic integrity, then iteratively accumulates sentences until exceeding predefined token limits (256 tokens); \textbf{(2) Semantic Chunking} employs SemanticChunker~\cite{langchain2024semanticchunker} to perform embedding similarity-based chunking using sentence embeddings to identify natural breakpoints in the text. The breakpoint threshold type is set to "percentile" and the breakpoint threshold amount is set to 50.0; \textbf{(3) Meta-Chunking}~\cite{DBLP:journals/corr/abs-2410-12788} employs perplexity-based semantic boundary detection with two sub-methods: PPL Chunking and Margin Sampling, utilizing Qwen/Qwen2.5-1.5B-Instruct model \cite{qwen2} for text chunking. PPL Chunking adopts the threshold of 0.5 as recommended in the original paper, while Margin Sampling uses default settings; \textbf{(4) LumberChunker}~\cite{DBLP:conf/emnlp/DuarteMGF0O24} provides LLM-guided chunk boundary prediction, utilizing the same configuration as the Generative Model.

\paragraph{\emph{FreeChunker} Setting}
The Sentenizer module adopts \textbf{Traditional Chunking} as the foundational sentence splitting method, with the same setting. For the granularity configuration, the simplest combination of three granularities is directly preset: (1, 2, 4). This setup ensures efficient processing while providing multi-scale context coverage.

\subsection{Main Results}

\paragraph{Overall Performance}
Table~\ref{tab:longbench_chunking} presents the comparative results on LongBench V2 across three embedding models. \emph{FreeChunker} achieves the best average performance among all methods.

\paragraph{Ablation Study}
Since \emph{FreeChunker} utilizes Traditional Chunking as the underlying chunking method for its Sentenizer to segment basic sentence units, and subsequently applies the Cross-Granularity Encoder to obtain cross-granularity embeddings, the performance gap between them can be regarded as an ablation study for the proposed cross-granularity framework. As shown in Table~\ref{tab:longbench_chunking}, \emph{FreeChunker} consistently outperforms the Traditional baseline, which verifies the superposition effect of multi-granularity embeddings and the effectiveness of the cross-granularity framework.

\paragraph{Time Efficiency}
Figure~\ref{fig:time_efficiency} illustrates the time cost distribution across different tasks. Compared to Traditional Chunking, \emph{FreeChunker} merely adds the forward propagation encoding time of the Cross-Granularity Encoder, resulting in negligible extra overhead. In contrast, LLM-based methods (e.g., LumberChunker) consume hundreds of seconds. This achieves a speedup of up to $30\times$ compared to complex semantic chunkers, demonstrating that \emph{FreeChunker} effectively circumvents the efficiency bottleneck of semantic-aware chunking without compromising retrieval quality.
\section{Conclusion}

This paper introduces \emph{FreeChunker}, a novel cross-granularity chunking framework that fundamentally transforms the traditional text chunking paradigm. By shifting from ``static boundary identification and re-encoding'' to ``flexible semantic combination of sentence encodings,'' this framework significantly reduces the computational overhead of semantic analysis required for boundary detection, while enhancing adaptability to diverse query requirements. Experimental results obtained using the simplest preset chunk patterns demonstrate significant advantages over existing methods, thereby validating the potential of this framework.

Beyond the preset chunk patterns explored in the current study, the framework's chunk pattern control mechanism, $\mathbf{P}_{g,s}[i,j]$, provides non-contiguous sentence linking capabilities. By supporting sophisticated combination methods based on sentence encodings, the framework enables the construction of complex non-adjacent sentence connections and patterns tailored to specific scenarios and tasks. These capabilities offer promising prospects for future research into adaptive and structural document modeling.

\section*{Limitations}

Although \textbf{\emph{FreeChunker}} requires increased GPU memory (VRAM) for encoding due to the introduction of the Cross-Granularity Encoder, this overhead is manageable. Since the encoding operates at the sentence level, the input sequence length for the attention mechanism is significantly shorter compared to token-level approaches, resulting in efficient computation.

\section*{Ethics Statement}

This work does not involve human subjects, animal experiments, or the collection of private or sensitive data. All datasets utilized in this research are publicly available and used in accordance with their respective licenses.

During the preparation of this manuscript, the authors used AI-based tools for grammatical refinement and stylistic polishing to improve the clarity of the presentation. Following this process, the authors conducted a thorough manual review and edit of the content to ensure its technical accuracy and original intent. The authors take full responsibility for the final content of the paper.

% \section*{Acknowledgements}

\bibliography{icml}
\bibliographystyle{acl_natbib}
\newpage
\appendix
% \input{Sections/appendix}
% \label{sec:appendix}

\end{document}